\pdfoutput=1

\pdfoutput=1

\documentclass[11pt]{article}

\usepackage{acl}

\usepackage{times}
\usepackage{latexsym}

\usepackage[T1]{fontenc}

\usepackage[utf8]{inputenc}

\usepackage{microtype}

\usepackage{graphicx}
\usepackage{listings}

\title{Parmesan: mathematical concept extraction for education}

\author{
  Jacob Collard \\
  NIST \\
  \texttt{jacob.collard@nist.gov} \\\And
  Valeria de Paiva \\
  Topos Institute \\
  \texttt{valeria@topos.institute} \\\AND
  Eswaran Subrahmanian \\
  Carnegie Mellon University \\
  \texttt{es3e@andrew.cmu.edu}
}

\begin{document}

\maketitle


\begin{abstract}

Mathematics is a highly specialized domain with its own unique set of challenges that has seen limited study in natural language processing. 
However, mathematics is used in a wide variety of fields and multidisciplinary research in many different domains often relies on an understanding of mathematical concepts. 
To aid researchers coming from other fields, we develop a prototype system for searching for and defining mathematical concepts in context, focusing on the field of category theory. 
This system Parmesan depends on natural language processing components including concept extraction, relation extraction, definition extraction, and entity linking.
In developing this system, we show that existing techniques cannot be applied directly to the category theory domain, and suggest hybrid techniques that do perform well, though we expect the system to evolve over time.
We also provide two cleaned mathematical corpora that power the prototype system, which are based on journal articles and wiki pages, respectively. 
The corpora have been annotated with dependency trees, lemmas, and part-of-speech tags.

\end{abstract}

\section{Introduction}
\label{sec:introduction}

Students and researchers in science and mathematics are often tasked with learning information about new fields of study for which relatively little pedagogical material is available.
Often, the student will have advanced education in one area, but require an introduction to a related multidisciplinary field in order to take advantage of advances made in an area different to their own.
For example, a physicist may need to learn about a new branch of mathematics that has recently been adapted to their field, or a computer scientist may need to learn about a new domain in order to develop an efficient algorithm to tackle a problem.
Because the student may be entering a new, highly advanced, and cutting edge area of study, simple pedagogical materials for that area don't necessarily exist yet.
At the same time, the student may lack the background to sufficiently understand the latest scientific articles and papers on the subject without additional help. 
Though mentors in the target area may be available in some cases, the increasing prevalence of multidisciplinary research across advanced fields means that some researchers may need to adapt quickly without much personal assistance.

In cases like this, machine-assisted learning has the potential to provide significant improvements to a learner's environment.
Several tasks common in the study of natural language processing (NLP) are relevant to this type of learning:

\begin{itemize}
    \item Definition extraction (DE) allows the learner to quickly find definitions for unfamiliar vocabulary.
    \item Entity linking (EL) can help a learner find new concepts in an established knowledge base.
    \item Relation extraction (RE) can help the learner make connections between unfamiliar concepts. 
    \item Collocation retrieval (CR) provides real-world context to new concepts.
\end{itemize}

The application to human learning provides additional context and constraints on these tasks, which are not always apparent in the academic discussion of NLP models.
Firstly, there may not be sufficient training data available in the target domain.
While a large language model (LLM) could be fine-tuned on the target vocabulary, there is likely limited data to fine-tune a model on the target tasks: explicitly annotated definitions, linked entities, and relations may occasionally be present in scientific articles, but may not be consistently annotated and may only appear sporadically. 

Secondly, the notion of a relevant concept needs to be considered from the perspective of a human learner. 
Specifically, a relevant concept is one which the human learner requires more information about. 
The system may not be able to predict which concepts are likely to require additional information, and this will vary from learner to learner in any case. 
Ideally, the system would be able to provide information about any concept that the user inquires about; a static set of concepts extracted from a corpus may not be sufficient.

Entity linking is both helped and hindered by the more specific context of this application.
Because a specific domain is known, it may be possible to filter candidate links based on information available in the knowledge base.
However, this can introduce difficulties for recall if relevant concepts are unnecessarily removed.
It also requires additional machinery to perform the filtering. 

Collocation extraction in this context is also unique.
A learner hopes to find exemplars of an unfamiliar concept in context.
These exemplars should appear in a variety of different contexts, highlighting possible different uses of the term.

While research has made great advances in domain adaptation, it is not clear that these advances will prove useful for the context of machine-assisted learning that we have described here.
Domain adaptation models perform best when they are given some training data in the target domain.
As mentioned above, some data may be available, but due to the high level of the target domains, annotations are likely to be sparse and corpora will generally be small. 
The specific constraints on tasks in this context also make domain adaptation more difficult. 
In Section \ref{sec:experiments}, we discuss some experiments that show that existing models do not perform well on the target data, even given opportunity for adaptation, though evaluation is difficult.

In this paper, we consider the case study of category theory (CT) as a domain.
Category theory is a well-established field, having existed since the 1940s, but it is still relatively small.
Because CT is focused on the mathematics of relations and composition, it has recently been applied to several different domains, including epidemiology \citep{2023-libkind}, chemistry \citep{baez2022}, and computer science \citep{moggi1991}.
This interdisciplinary growth makes CT an ideal candidate for this case study.

Category theory also poses several unique challenges.
While some of these are specific to category theory, similar challenges are likely to be present in other domains as well.

\begin{itemize}
    \item Many technical terms are identical to common English words, such as `category', `limit', `group', `object', and `ring'.
    \item Other technical terms do not appear in everyday English at all, such as `monad', `groupoid', or `colimit'. 
    \item Special symbols and diagrams are often dispersed through the text.
    \item Abbreviations and shortcuts are common.
\end{itemize}

Our goal is to present and analyze a prototype system, called Parmesan, that handles  entity linking and collocation search for category theory, from the perspective of a human learner.
In Section \ref{sec:corpus}, we discuss the development of corpora for the specific case study of category theory, and discuss operations to prepare these corpora for later use. 
The corpora are made freely available online, and can be applied to tasks other than those that we discuss here.
In Section \ref{sec:experiments}, we reproduce several experiments showing that simply transferring existing NLP technologies is not sufficient to provide the system that we want, though evaluation poses a significant challenge.
In Section \ref{sec:entity-linking}, we discuss methods for providing entity linking in the prototype system  Parmesan\footnote{A demo is available at \url{http://jacobcollard.com/parmesan2}}.
In Section \ref{sec:search}, we add collocation search to Parmesan.
We discuss the future work necessary to further improve Parmesan in Section \ref{sec:discussion}, including the addition of definition and relation extraction tools to the system.

\section{Corpus Preparation}
\label{sec:corpus}

We prepare two initial mathematical corpora for use in the study of mathematical language processing.
The first corpus consists of 755 abstracts (3188 sentences) from \emph{Theory and Applications of Categories} (TAC), a journal of category theory.
This corpus is very similar to the one presented in \citet{collard-etal-2022-extracting}, but has undergone additional processing and cleaning, which we describe here.
This corpus was selected as an exemplar for state-of-the-art mathematical research.
Many new concepts will be introduced in these abstracts, and more fundamental concepts will be used in practical contexts.

The second corpus consists of 11653  articles (175151 sentences) from the online encyclopedic resource for Category Theory nLab\footnote{http://ncatlab.org}. 
This corpus has undergone similar preprocessing to TAC.
It was selected as an exemplar for fundamental concepts in category theory and as a basic reference.
In addition to mapping concepts directly to nLab articles, it is also possible to see concepts used in the context of other articles. 
For example, in addition to the article on categories itself, the word ``category'' appears in many other contexts within nLab that can help to elucidate its meaning.
To prepare this corpus for use, we remove the Markdown markup, leaving only plain text. 
We have also filtered out documents describing books as well as meta-articles such as lists and categories.

To handle \LaTeX{} markup in both TAC and nLab, we use the LaTeXML converter\footnote{https://dlmf.nist.gov/LaTeXML} to identify mathematical expressions. 
Completely removing mathematical expressions could introduce problems; since we later apply parsing to the corpora, the gaps caused by removing inline math will produce ungrammatical sentences and thus invalid dependency trees.
However, in their raw form, mathematical expressions are represented using \LaTeX{}, which can be difficult to read, especially when used to represent complex formulas.
Therefore, we convert mathematical expressions into plain text phrases that approximate the original mathematical formulas, providing the parser with linguistic material to work with without the structure being overly saturated with markup.

Though these corpora do not include large amounts of annotation, they are associated with some useful metadata.
The TAC corpus includes titles, authors, dates, and keywords selected by the authors to describe their abstracts. 
These keywords will be used as part of the evaluation in the next section.
The nLab corpus includes titles and dates, of which the titles are used as part of the evaluation in the next section.

For both corpora, we also provide annotations of dependency trees, part of speech tags, and lemmas in CONLL-U format\footnote{https://universaldependencies.org/format.html}. 
These annotations were generated automatically using the open NLP framework spaCy\footnote{http://spacy.io}.

\section{Experiments}
\label{sec:experiments}

Research on entity linking, definition extraction, and relation extraction has covered many different domains with a variety of methods.
In this section, we describe previous research on these topics, including their results in different domains related to mathematics, as well as our own experiments which evaluate these methods in the domain of mathematics.
There is relatively little research applied directly to the domain of mathematics, so we are often forced to study related domains such as sciences, which face some of the same challenges as mathematics. 
Where possible, we also attempt to reproduce the results of previous work, using our own corpora. 
Unfortunately, rigorous evaluation is not always possible, due to a lack of annotated mathematical data. 

\subsection{Entity linking}
\label{subsec:entity-linking}

Entity linking is an NLP task in which a system attempts to identify important concepts in a corpus and link them to a knowledge base, such as Wikidata\footnote{\url{wikidata.org}}.
The task can often be divided into two parts: candidate identification and linking. 
The candidate identification subtask is similar to named entity recognition, although the notion of what constitutes an entity is somewhat different.
Linking is similar in some ways to word sense disambiguation, in that the correct knowledge base record must be identified in the case that a word or phrase cannot be unambiguously attached to a single record. 
By linking entities to a knowledge base, a system can provide information about entities in a corpus from a manually curated repository such as WikiData.

One benchmark for entity extraction is the SciERC dataset \citep{luan2018multitask}, which is a collection of 500 scientific abstracts.
The dataset includes human annotations for six types of scientific entities and seven types of relation. 
Since the sciences often make use of mathematical notations and concepts, this domain has some similarities to the domain of mathematics. 

Several methods have been evaluated against the SciERC dataset. 
These results are summarized in Table \ref{tab:entity-extraction-scierc}.
As shown, the top performing models are about 70\% accurate for entity extraction, but only about 50\% accurate for relation extraction on the SciERC benchmark. We describe these methods below.

\begin{table}
    \centering
    \begin{tabular}{|c|c|c|}
        \hline
        \textbf{Model} & \textbf{Entities} & \textbf{Relations} \\
        \hline
        PL-Marker & 0.699 & 0.532 \\
        SpERT.PL & 0.7053 & 0.5125 \\
        SpERT & 0.703 & 0.5084 \\
        Cross-Sentence & 0.681 & 0.501 \\
        DyGIE++ & 0.675 & 0.484 \\
        \hline
    \end{tabular}
    \caption{$F_1$ scores for several models on the SciERC benchmark}
    \label{tab:entity-extraction-scierc}
\end{table}

We have reproduced the entity extraction component for some of these models on our  TAC corpus.
The models are not provided additional training, since large amounts of training data are not available.
However, we do have some data to evaluate the results of these models on the TAC corpus.

Specifically, we have two sets of silver standard keywords available for this corpus.
The first set of keywords is based on the keywords selected by the authors of the TAC abstracts. 
Since the author-selected keywords do not necessarily occur in the actual text of the corpus, we filter out any keywords which do not appear in the text, since the key phrase extraction algorithms we evaluate are only capable of extracting keywords that appear in the text.
The second set of keywords consists of titles of articles in the nLab corpus, which are filtered to discount non-extractive keywords.
This identifies likely category theory concepts which appear in TAC.
Since nLab is effectively an encyclopedia of category theory, most of its articles describe concepts in category theory, though a small percentage of articles about non-conceptual entities such as books, people, and metadata also exist. We have done our best to filter these out, as described in Section \ref{sec:corpus}. 

In Table \ref{tab:terminology-extraction-author}, we show the results of five terminology extraction methods when evaluated against TAC author keywords.
In Table \ref{tab:terminology-extraction-titles}, we show the results of the same terminology extraction methods evaluated against nLab titles appearing in TAC.

\begin{table}
    \centering
    \begin{tabular}{|c|c|c|c|}
        \hline 
        \textbf{Model} & \textbf{Precision} & \textbf{Recall} & \textbf{$F_1$} \\
        \hline
        DyGIE++ & 0.22 & 0.35 & 0.27 \\
        MWE & 0.12 & 0.78 & 0.20 \\
        PL-Marker & 0.23 & 0.38 & 0.28 \\
        SpERT.PL & 0.14 & 0.77 & 0.23 \\
        Textrank & 0.15 & 0.55 & 0.23 \\
        \hline
    \end{tabular}
    \caption{Terminology extraction algorithms evaluated against TAC author keywords}
    \label{tab:terminology-extraction-author}
\end{table}

\begin{table}
    \centering
    \begin{tabular}{|c|c|c|c|}
        \hline
        \textbf{Model} & \textbf{Precision} & \textbf{Recall} & \textbf{$F_1$} \\
        \hline
        DyGIE++ & 0.12 & 0.27 & 0.16 \\
        MWE & 0.08 & 0.75 & 0.14 \\
        PL-Marker & 0.11 & 0.27 & 0.16 \\
        SpERT.PL & 0.08 & 0.63 & 0.14 \\
        Textrank & 0.09 & 0.46 & 0.14 \\
        \hline
    \end{tabular}
\caption{Terminology extraction algorithms evaluated against nLab titles appearing in TAC}
    \label{tab:terminology-extraction-titles}
\end{table}

The models that we evaluate include:

\begin{itemize}
    \item \textbf{DyGIE++}: a joint entity-, relation-, and event-extraction model using transformers \citep{Wadden2019EntityRA}.
    \item \textbf{MWE}: single- and multi-word noun phrases extracted using mwetoolkit3 \citep{ramisch2012mwe}. The exact patterns used can be found in our freely available code.
    \item \textbf{PL-Marker}: a neural entity- and relation-extraction model which models relations between spans using a novel packing strategy \citep{ye2022plmarker}.
    \item \textbf{SpERT.PL}: a joint entity- and relation-extraction model which incorporates linguistic information such as part-of-speech tags \citep{saietal2021spert}.
    \item \textbf{Textrank}: a Python implementation\footnote{https://github.com/DerwenAI/pytextrank} of the graph-based textrank algorithm \citep{mihalcea2004}.
\end{itemize}

Notably, the results for these models on the TAC corpus (using both author keywords and nLab titles) are significantly lower than they were for SciERC.
There are several possible factors contributing to this.
First, these models have not been specifically trained on mathematical texts, though some models (MWE and Textrank) do not benefit from training in any case.
Mathematical texts likely have many features which are not common in scientific ones, such as formulas, variables, and unique vocabulary.

Secondly, the types of entities annotated in SciERC are quite different from the kinds selected by our silver standards.
Specifically, the entity types do not all apply to mathematics in general.
SciERC entities are labeled with one of the following tags: Task, Method, Metric, Material, Other-ScientificTerm, and Generic \citep{luan2018multitask}. 
While methods and metrics may be used in mathematics, they are uncommon in category theory, and materials are rare across the board.
The definition of Task in SciERC includes problems to solve and systems to construct, so it likely has some overlap with math.
However, there are classes of mathematical terms that are not represented well by this annotation scheme. 
For example, theorems cannot be described, except under the broad heading of Other-ScientificTerm.
More importantly, abstract concepts like "set" and "category" are not well-described by this system.

The differences between the author keyword evaluation and the nLab title evaluation is also of interest.
All five models performed better against author keywords compared to nLab titles, in both precision and recall. 
All models also performed better in recall than in precision across the board. 
This may be in part because neither silver standard captures all possible mathematical terms in the corpus, so some of the predicted keywords may not be represented in the evaluation, even if they are valid.
However, for our use case, recall is significantly more important than precision, since we hope to have results for as many possible learner queries as possible.

Only two models, MWE and SpERT.PL were able to achieve recall above 0.7 on the author keywords, and only MWE was able to achieve recall above 0.7 on nLab titles. 
DyGIE++ and PL-Marker both perform poorly compared to their accuracy on SciERC. 

Clearly, these models do not directly transfer to the mathematical domain without additional fine-tuning.
Unfortunately, a lack of adequate training data makes this difficult. 
Neural methods in particular seem more sensitive to this transfer than others, such as Textrank and MWE, though these have their own weaknesses, such as poor precision. 

Once entity extraction is complete, the next stage of entity linking is to link each extracted entity to a knowledge base.
In our use case, entity linking is the main goal, and entity extraction can take something of a smaller role.
When the user enters search queries, it can be assumed that the user has input a valid query, since the user has provided it for us.
The entities that the user provides still need to be linked to a knowledge base, and entity extraction  plays a role in evaluation.

Entity linking models have been evaluated against several different benchmarks, including AIDA\citep{mpi-aida}, WiC-TSV \citep{breit-etal-2021-wic}, and many others.
There has also been significant research on the area of mathematical entity linking in particular, though this research tends to focus on formulas, while our goal is to link entities found primarily in text. 
Some models, however, do make use of both text and formulas, and have been evaluated against datasets such as Wikipedia with the goal of linking to data in Wikidata \citep{scharpf-et-al-fast-linking}. 
Despite focusing on mathematical entity linking, however, the domain that this model has been evaluated against is actually physics, which makes significant use of mathematical concepts and formulas, but is distinct from the mathematical domain we are considering, as shown by the differences in score between SciERC and TAC in earlier experiments.

\subsection{Definition Extraction}
\label{subsec:definition-extraction}

There are often two components or strategies involved in definition extraction: identifying sentences that contain definitions, and identifying the terms and definitions precisely in text. Some models, such as \citep{ben_veyseh_joint_2019}, use a joint model which simultaneously identifies definitional sentences and precise terms and definitions.
This model is evaluated against three datasets: WCL, WC00, and DEFT. 
Word Class Lattices (WCL) is a definition extraction benchmark consisting of sentences from Wikipedia which distinguishes between definitional and non-definitional sentences \citep{navigli-velardi-2010-learning}.
WC00 also distinguishes between definitional and non-definitional sentences, and contains over 2000 sentences from the ACL anthology from the scientific domain \citep{jin-etal-2013-mining}.
DEFT consists of two subcorpora: one covering textbooks from domains including biology, history, and physics; and one covering contracts \citep{spala-etal-2019-deft}. 

This joint model described in \citet{ben_veyseh_joint_2019} reports an $F_1$ score of up to 85.3 on WCL, 66.9 on W00, 54.0 on DEFT Textbooks, and 71.7 on DEFT Contracts.
Another model, \citep{vanetik-etal-2020-automated} is designed specifically for mathematics and combines dependency tree and word vector representations to construct a neural classification model.
Their model scores above 0.9 on WCL and above 0.82 on WC00, and above 0.8 on WFM, a benchmark drawn from Wolfram MathWorld specifically for mathematics \citep{vanetik-etal-2019-wfm}
This shows significant variation between different domains.

For our category theory domain, annotated definition data is not available to evaluate these systems.
Given that \citet{vanetik-etal-2020-automated} is evaluated on the mathematical WFM corpus, we expect it to perform reasonably well on our data.
Other models, such as \citet{ben_veyseh_joint_2019}, are trained and evaluated on scientific corpora, such as DEFT Textbooks, so we expect some overlap, but also some variation.

A qualitative evaluation will be the subject of future work.

\section{Entity Linking with Parmesan}
\label{sec:entity-linking}

The goal of the Parmesan system is to allow a learner in the domain of category theory to find information and context about novel concepts.
In most cases, there is little need for direct concept extraction: the user provides a query which describes the concept of interest. 
For example, if the user searches for ``double category'', there is no need to check whether ``double category'' can be extracted directly, since this is clearly a concept of interest for the user.
The more important challenge for Parmesan is to provide information about whatever the user searches.
In the case of entity linking, this means providing links to knowledge bases which describe the user's term.

One challenge is that the user generally provides a term without any specific context.
This makes the use of neural entity linking systems more difficult, though the problem is mitigated by the known domain of use.
Since the system is intended specifically for category theory, the entity linking system can focus on links that are known to describe category theoretic concepts, and filter out concepts that are unlikely to be mathematical in nature.

We use Wikidata as our first initial knowledge base.
While many categorical concepts are still missing from Wikidata, it is relatively complete and well-established.
Wikidata also provides several tools to access its data that allow filtering entities by their properties and relations.
One of the most powerful of these is the Wikidata Query Service\footnote{https://query.wikidata.org/} which provides an API for queries using the SPARQL Protocol and RDF Query Language (SPARQL). 

A simple SPARQL query over Wikidata can identify concepts that have a given label or alias.
For example, to search for entries for ``category'', we can use the query in Listing \ref{lst:initial-sparql} to identify Wikidata items, their labels, and their descriptions. This will match any entity which either has the label or alias ``category'', which includes several entities, summarized in Table \ref{tab:initial-query-results}. 

\begin{lstlisting}[captionpos=b, caption=Unfiltered SPARQL query, label=lst:initial-sparql, basicstyle=\ttfamily\scriptsize,frame=single]
SELECT DISTINCT ?item ?itemLabel ?itemDescription
WHERE {
  {?item rdfs:label "category"@en.} UNION
  {?item skos:altLabel "category"@en.}
  SERVICE wikibase:label {
    bd:serviceParam wikibase:language "en".
  }
}
\end{lstlisting}

\begin{table*}
    \centering
    \begin{tabular}{|l|p{11em}|p{20em}|}
        \hline
        Entity & Label & Description \\
        \hline
        Q15846545 & category & class of sets \\
        \hline
        P910 & topic's main category & main Wikimedia category \\
        \hline
        Q4167836 & Wikimedia category & use with 'instance of' (P31) for Wikimedia category \\
        \hline
        Q21146257 & type & kind or variety of something \\
        \hline
        Q15013692 & Category:Rhaba & Wikimedia category \\
        \hline
        Q9757078 & category & Wikimedia category \\
        \hline
        Q719395 & category & algebraic structure of objects and morphisms between objects, which can be associatively composed if the (co)domains agree \\
        \hline
        Q4118499 & category & in Kantian philosophy, a pure concept of the understanding (Verstand); a characteristic of the appearance of any object in general, before it has been experienced \\
        \hline
        Q16781549 & Category:Biographical plays about English royalty	& Wikimedia category \\
        \hline
        Q64549097 & category & concept \\
        \hline
    \end{tabular}
    \caption{Results of the query in Listing \ref{lst:initial-sparql}}
    \label{tab:initial-query-results}
\end{table*}

Entity Q719395 is the entity that is most relevant to category theory, though Q15846545 is also relevant to mathematics. 
Some of these entries are clearly not relevant to the domain, as they are metadata (P910, Q4167836, Q15013692, Q16781549), while others are unlikely to be relevant (but may be in certain circumstances) because they are from different domains (such as Q15846545, Q21146257, Q4118499, Q64549097). 

Ideally, we would filter and sort the results to present the most likely category theory concepts first and either remove or deprioritize concepts that are less likely to be from category theory. 
Some of these are relatively straightforward to identify; others require access to information which is not readily available in Wikidata.
Some entries may also be of unclear utility.
For example, the difference between Q64549097 ``category'' and Q21146257 ``type'' is not clearly described, and Q64549097 has an unfortunately sparse entry. 
Since Wikidata is a large, crowdsourced knowledge base, there is significant room for inconsistencies in the database.

We have identified several entries that are unlikely to describe categorical concepts.
These are Wikidata entries whose children are well outside the scope of mathematics. 
The set of such entries is given in Table \ref{tab:removed-categories}. 

\begin{table}
    \centering
    \begin{tabular}{|c|c|}
        \hline
        ID & Description \\
        \hline
        Q223557 & Physical object \\
        Q4167836 & Concrete object \\
        Q17334923 & Physical location \\
        Q4167836 & Wikimedia category \\
        Q1914636 & Activity \\
        Q3769299 & Human Behavior \\
        Q63539947 & Artistic Concept \\
        Q186408 & Point in Time \\
        Q186081 & Time Interval \\
        Q8142 & Currency \\
        \hline
    \end{tabular}
    \caption{Classes of Wikidata entry to filter out}
    \label{tab:removed-categories}
\end{table}

Some of these classes may be tangentially related to mathematics in certain contexts, e.g., in applications of category theory to natural and social sciences. 
However, since our focus is on mathematics, we assume that users are interested in the mathematical uses of terms and not in concepts from other fields described in mathematics.

Though ordering concepts related to mathematics first may allow for greater recall without sacrificing precision, for now we simply filter out concepts matching any of the above classes. 
This can be done by refining Listing \ref{lst:initial-sparql} with additional filters. 
Ideally, these filters would remove any subclass of the given categories; however, for performance reasons, it only removes immediate subclasses.

\begin{lstlisting}[captionpos=b, caption=Unfiltered SPARQL query, label=lst:final-sparql, basicstyle=\ttfamily\scriptsize,frame=single]
SELECT DISTINCT ?item ?itemLabel ?itemDescription 
WHERE {
  {?item rdfs:label "category"@en.} UNION
  {?item skos:altLabel "category"@en.}
  MINUS { ?item wdt:P279 wd:Q223557 }
  MINUS { ?item wdt:P279 wd:Q4167836 }
  MINUS { ?item wdt:P279 wd:Q17334923 }
  MINUS { ?item wdt:P279 wd:Q4167836 }
  MINUS { ?item wdt:P279 wd:Q1914636 }
  MINUS { ?item wdt:P279 wd:Q3769299 }
  MINUS { ?item wdt:P279 wd:Q63539947 }
  MINUS { ?item wdt:P279 wd:Q186408 }
  MINUS { ?item wdt:P279 wd:Q186081 }
  MINUS { ?item wdt:P279 wd:Q8142 }
  SERVICE wikibase:label {
    bd:serviceParam wikibase:language "en".
  }
}
\end{lstlisting}


\section{Search with Parmesan}
\label{sec:search}

\begin{figure*}[ht]
    \centering   
    \includegraphics[width=\textwidth]{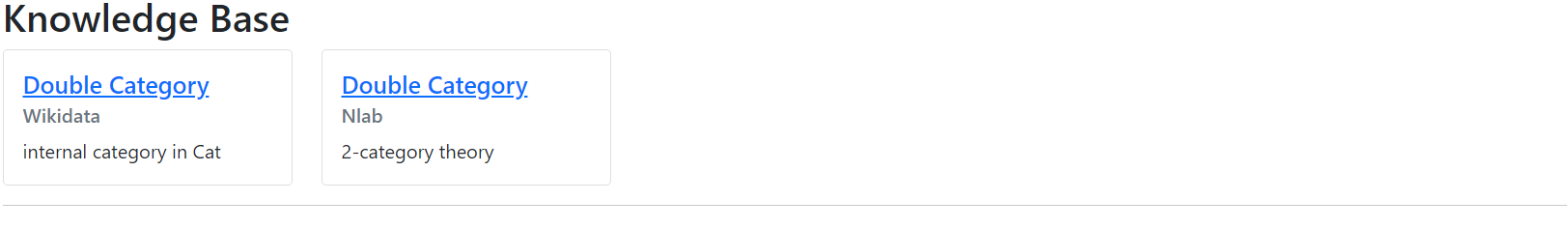}
    \includegraphics[width=\textwidth]{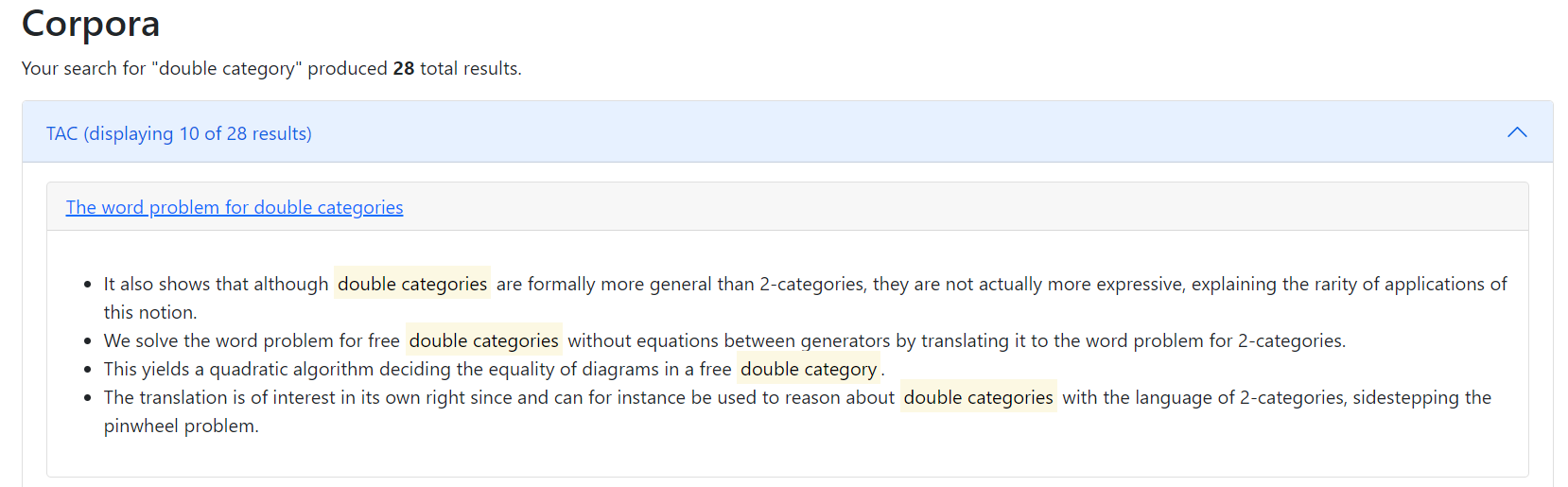}
    \includegraphics[width=\textwidth]{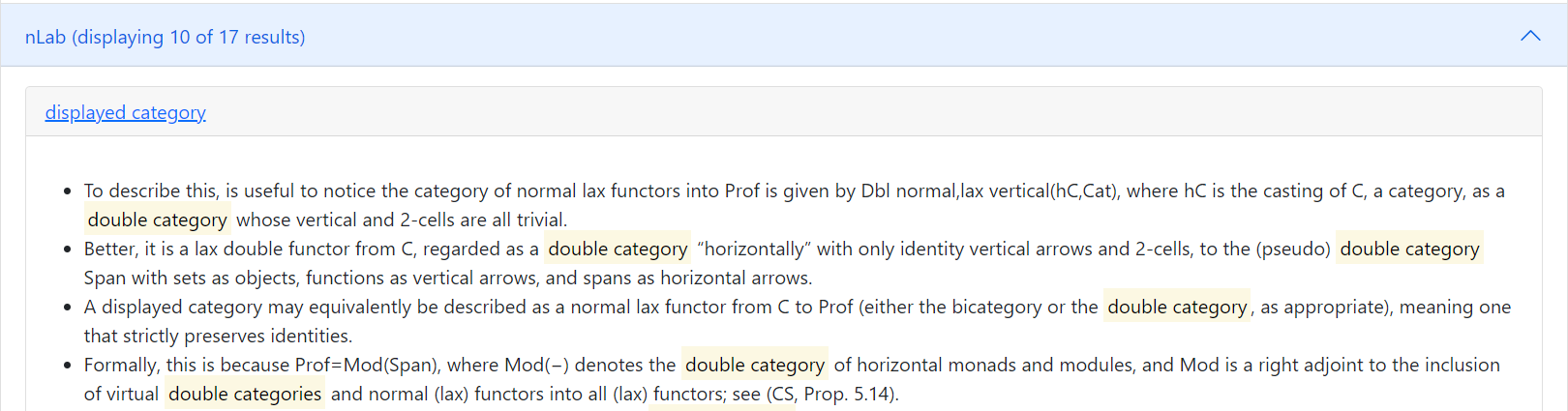}
    \caption{An example of Parmesan search results}
    \label{fig:search-results}
\end{figure*}

In addition to entity linking with Wikidata and nLab, Parmesan also provides text search over data in the TAC and nLab corpora.
This search is intended to allow users to identify the contexts in which an unfamiliar term appears. 
This complements the definitions provided by entity linking by showing how the terms are actually used in different mathematical contexts.

There are two types of context that Parmesan provides.
At a high level, the two corpora (TAC and nLab) represent two broad categories.
The TAC corpus, being drawn from journal articles, provides a view into the state-of-the-art, advanced concepts, and newly-coined phrases.
The nLab corpus, on the other hand, being a community wiki, is primarily dedicated to descriptions of common, high-level concepts in category theory.
Thus, this corpus provides contexts about the relationships between concepts and their definitions.
Though there are still many advanced concepts in this corpus, they are presented in a more approachable style and at a more general level.

Both corpora also provide precise contexts in the form of exemplars of sentences and phrases where the target term is used.
We can easily identify matches of the phrase that the user has input by finding corresponding lemmas in the annotated corpus.
We use lemmas to show the term used with different inflections.
We then display all sentences that contain the corresponding lemmas.
Currently, the sentences are returned in an arbitrary order. 
However, the distinction between the two corpora is kept clear: results from TAC are returned separately from nLab.
This allows the learner to clearly distinguish between different contexts in which terms appear, both at the sentential level, at the document level, and at the corpus level.
Links are provided to individual nLab articles and to specific TAC abstracts if the user requires more information or additional context.

Figure \ref{fig:search-results} shows an example of search results found by Parmesan for the search term ``double category''.
At the top is the list of knowledge base entries found for that concept; in this case, there is  the concept of a double category from category theory in WikiData (Wikidata entry Q99613675) and the nLab entry on double categories. 

Next are shown results from TAC and nLab sentences.
Each document is displayed as a separate card with a link to the original document (TAC abstract or nLab article). 
A list of sentences containing the search term are then shown within the card.
The search term is highlighted where it appears in the text of each sentence.
As can be seen by this example, variant forms of the word (such as the plural ``categories'') are shown as well as the exact terms searched by the user.
However, there is no additional semantic or vector-based search to identify similar concepts to ``double'' or ``category''. 
Since the system is aimed primarily at learners, we hope to keep the analysis straightforward and easily interpretable to the user. 

The example sentences in Figure \ref{fig:search-results} reveal certain facts about double categories that are useful to a newcomer in the field:

\begin{itemize}
    \item They are formally more general than 2-categories.
    \item There is a subtype of double category called a free double category.
    \item There are mathematical problems of interest for double categories, such as the word problem for free double categories.
\end{itemize}

This is largely cutting-edge information about double categories; more general and approachable information about them may be found in nLab, which is displayed separately. 

The user is also able to hide and display the TAC and/or nLab corpora individually if they are only searching for information from a certain set of contexts.

\section{Discussion}
\label{sec:discussion}

The current implementation of Parmesan provides a tool for learners and researchers in the field of category theory to search for concepts to find their contexts of use and information about them in Wikidata.
This provides the user with different points of view about a concept:

\begin{itemize}
    \item concise but highly structured, interconnected data in Wikidata;
    \item the expert, but general and pedagogical, view of nLab; and
    \item the cutting-edge research point-of-view of TAC.
\end{itemize}

Each of these points of view may be useful to different users, and separating them in the display allows the user to compare and contrast different contexts of use of the words they are looking for, providing real-world examples and practical information about novel concepts.

This style of interactive learning can be further improved as we incorporate resources from other sources and new natural language processing methods.
For example, new corpora can be incorporated to provide new contexts of use for concepts. 
For example, adding a repository of articles from a category theory subsection of arXiv would add contexts from new preprints and a broader class of mathematical journal articles.
Similarly, we can incorporate entity linking to other databases such as Planet Math\footnote{\url{https://planetmath.org}} or the Encyclopedia of Mathematics\footnote{\url{https://encyclopediaofmath.org/wiki/Main_Page}}.

We can also incorporate new advances in natural language processing and technology.
As shown in Section \ref{sec:experiments}, relation extraction suffers from challenges in specific domains such as category theory.
Since the relation extraction algorithms we study are unable to accurately extract mathematical concepts, the relations that build on these concepts are generally lacking as well.
With additional training or other advances in relation extraction, the addition of relations to the interface would introduce a new type of context to users.
By understanding how concepts are related to one another, a learner can understand the meaning of that concept in terms of more familiar ideas.
Adding definition extraction, semantic similarity search, and other natural language processing methods to the system can grant it similar improvements.

Other future work for the system includes improving the order of search results, better filters on Wikidata links, and various performance improvements.
The addition of automatic definition extraction is also considered to be of particular importance, since definitions as they appear in context will be especially useful to learners. 

The principles of Parmesan are by no means limited to category theory, though category theory does pose some unique challenges and provides some unique opportunities due to its growing presence in interdisciplinary research.
Interfaces similar to Parmesan could, however, be applied to any field.

Overall, Parmesan provides a new approach to search for learners new to the field of category theory.
This approach is centered around providing context and domain-specific knowledge about user concepts.
Because the user provides the concepts, there is less need for error-prone concept extraction, and we can instead rely on entity linking, taking advantage of known properties of the domain.
The system provides several different contexts, allowing the user to compare and contrast disparate sources of knowledge to find the information they need about novel concepts.

\section{Disclaimer}
\label{sec:disclaimer}

Certain commercial entities, equipment, or materials may be identified in this document in order to describe an experimental procedure or concept adequately.
Such identification is not intended to imply recommendation or endorsement by the National Institute of Standards and Technology, nor is it intended to imply that the entities, materials, or equipment are necessarily the best available for the purpose.

\bibliography{main}
\bibliographystyle{acl_natbib}
\end{document}